\documentclass[sigconf]{acmart}
\AtBeginDocument{%
  \providecommand\BibTeX{{%
    \normalfont B\kern-0.5em{\scshape i\kern-0.25em b}\kern-0.8em\TeX}}}

\copyrightyear{2021}
\acmYear{2021}
\setcopyright{acmlicensed}\acmConference[SIGIR '21]{Proceedings of the 44th
International ACM SIGIR Conference on Research and Development in
Information Retrieval}{July 11--15, 2021}{Virtual Event, Canada}
\acmBooktitle{Proceedings of the 44th International ACM SIGIR Conference on
Research and Development in Information Retrieval (SIGIR '21), July 11--15,
2021, Virtual Event, Canada}
\acmPrice{15.00}
\acmDOI{10.1145/3404835.3463007}
\acmISBN{978-1-4503-8037-9/21/07}

\usepackage{multirow}
\usepackage{balance}

\begin{document}
\fancyhead{}
 
\title{GraphPAS: Parallel Architecture Search for Graph Neural Networks}

\author{Jiamin Chen}
\email{chenjiamin@csu.edu.cn}
\affiliation{
\institution{School of Computer Science and Engineering, Central South University}
}

\author{Jianliang Gao}
\email{gaojianliang@csu.edu.cn}
\affiliation{
\institution{School of Computer Science and Engineering, Central South University}
}
\authornote{Corresponding author.} 

\author{Yibo Chen}
\email{chenyibo8224@gmail.com}
\affiliation{
\institution{ State Grid Hunan Electric Power Company Limited}
}

\author{Oloulade Babatounde Moctard}
\email{oloulademoctard@csu.edu.cn}
\affiliation{
\institution{School of Computer Science and Engineering, Central South University}
}

\author{Tengfei Lyu}
\email{tengfeilyu@csu.edu.cn}
\affiliation{
\institution{School of Computer Science and Engineering, Central South University}
}

\author{Zhao Li}
\email{lizhao.lz@alibaba-inc.com}
\affiliation{
\institution{Alibaba Group}
}

\begin{abstract}
Graph neural architecture search has received a lot of attention as Graph Neural Networks (GNNs) has been successfully applied on the non-Euclidean data recently. However, exploring all possible GNNs architectures in the huge search space is too time-consuming or impossible for big graph data. In this paper, we propose a parallel graph architecture search ($GraphPAS$) framework for graph neural networks. In $GraphPAS$, we explore the search space in parallel by designing a sharing-based evolution learning, which can improve the search efficiency without losing the accuracy. Additionally, architecture information entropy is adopted dynamically for mutation selection probability, which can reduce space exploration. The experimental result shows that $GraphPAS$ outperforms state-of-art models with efficiency and accuracy simultaneously.
\end{abstract}


\maketitle

\section{Introduction}
\begin{figure}[t]
  \centering
  \includegraphics[width=\linewidth]{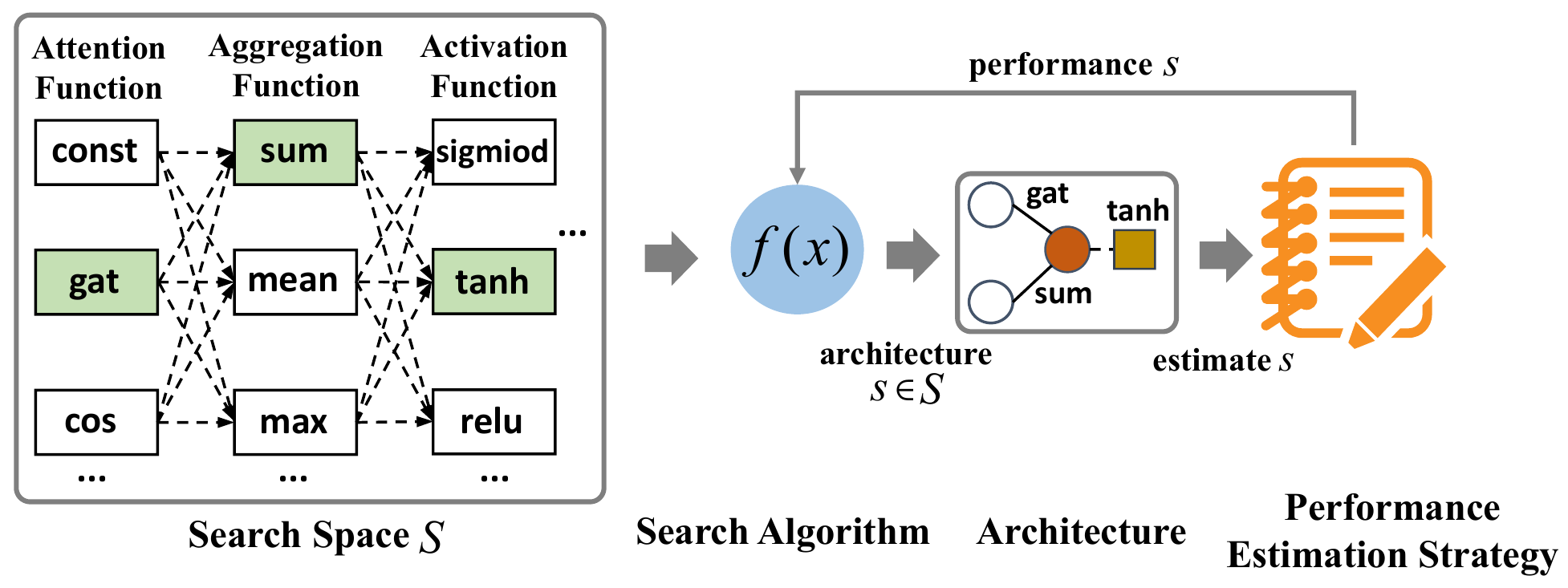}
  \caption{Illustration of graph neural architecture search.}
  \label{nas_framework}
  \Description{description.}
\end{figure}

In recent years, Graph Neural Networks (GNNs) have demonstrated great practical performance for analyzing graph-structured data \cite{yu2020tagnn}. GNNs usually consist of different components such as attention, aggregation, activation functions and so on. In traditional GNNs, the selection of components is a tough problem and determined by manual-based experience, which is difficult to achieve optimal results. Recently, neural architecture search (NAS) has attracted increasing research interests \cite{elsken2019neural}. 
NAS aims to explore the optimal neural architecture from search space for maximizing model performance. As shown in Fig. \ref{nas_framework}, there are a lot of components of GNNs architecture, such as attention, aggregation and activation function, which form the search space of graph neural networks. Architecture search algorithm samples a combination of components from search space $S$ such as $\{gat, sum, tanh,...\}$ in the example as a GNNs architecture $s$. The architecture $s$ is further evaluated by performance estimation strategy. The search algorithm uses the estimated performance to generate new and better GNNs architecture from search space $S$.
Recently, some works have focused on the problem of graph neural architecture search. GraphNAS \cite{gao2020graph} uses a recurrent network to generate fixed-length strings that describe the architectures of graph neural networks, and trains the recurrent network with policy gradient to maximize the expected accuracy of the generated architectures on a validation data set. Gene-GNN \cite{shi2020evolutionary} is proposed through the evolution approach to identify both the GNNs architecture and training hyper-parameter in two search spaces.

However, the main challenge of graph neural architecture search is the efficiency since the search space is explosively related to the graph neural network structure. For example, there is about $10^{8}$ possibility of combinations for two layers GNNs architecture \cite{gao2020graph}. Exploring all possible GNNs architectures in the huge search space is too time-consuming or impossible for big graph data. 
In this paper, we propose a parallel graph neural architecture search framework ($GraphPAS$) that can speed up the search with high accuracy simultaneously.
In the $GraphPAS$, we design sharing-based evolution learning to enhance the diversity of population. For the evolutionary search, we utilize information entropy to assess the mutation selection probability of architecture component, which can reduce exploration effectively in the huge search space for achieving good GNNs architecture. The main contributions of this paper are summarized as follows:
\begin{itemize}
\item To the best of our knowledge, we make the first attempt to parallel search framework for graph neural architecture search.  
\item We propose a novel neural architecture search framework $GraphPAS$ to find the optimal GNNs architecture automatically. In the $GraphPAS$, we design sharing-based evolution learning and adopt architecture information entropy to restrict search direction, which can improve the search efficiency without losing the accuracy .
\item The experiments shows that $GraphPAS$ outperforms state-of-art models in both efficiency and accuracy on real-world datasets.
\end{itemize}

\begin{figure*}[htbp]
  \centering
  \includegraphics[width=1\linewidth]{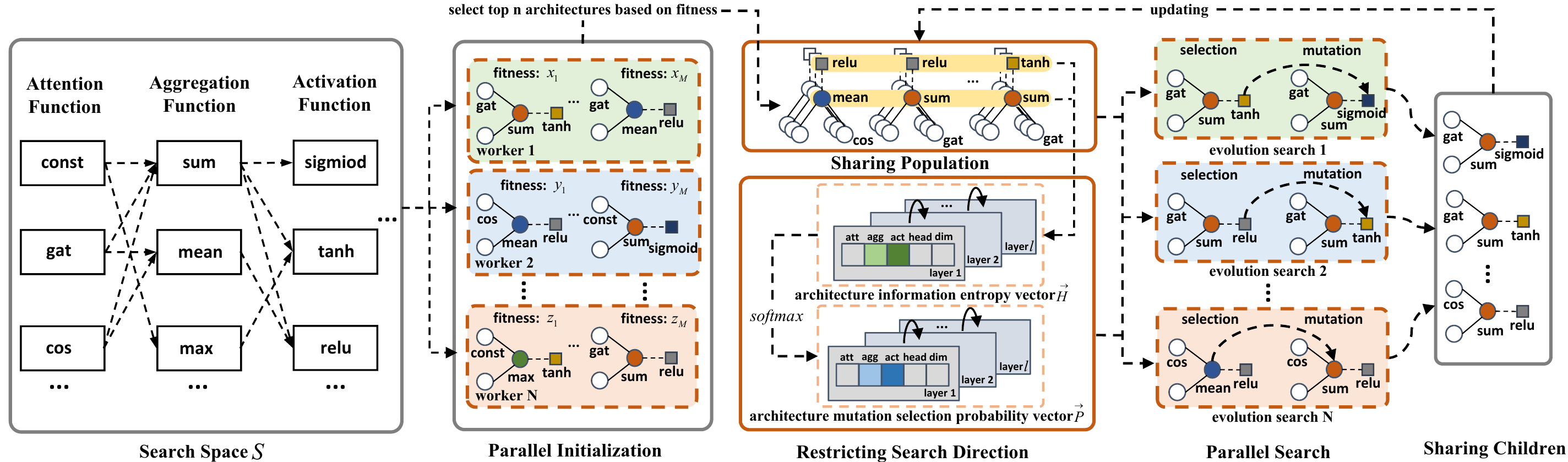}
  \caption{
Parallel search framework for graph neural architecture search.
First, $N \times M$ GNNs architectures are generated randomly by $N$ workers in parallel as initial population.
Then, selecting top $n$ GNNs architectures based on fitness as sharing population.
Subsequently, computing information entropy vector $\vec{H}$ based on frequency distribution of architecture component value in sharing population.
Next, utilizing vector $\vec{H}$ to dynamically identify mutation selection probability vector $\vec{P}$.
After, each worker samples child architecture simultaneously based on sharing population and architecture mutation selection probability vector $\vec{P}$.
Finally, merging child architectures generated by $N$ workers as sharing children, putting child architecture into sharing population if the fitness of child architecture is greater than threshold.}
\label{fram}
\Description{description.}
\end{figure*}

\section{The Proposed GraphPAS METHOD}
\label{method}
In this section, we first define the problem formulation of graph neural architecture search and then introduce the search space of GNNs. 
Finally, we discuss the parallel search algorithm of $GraphPAS$.

\subsection{Problem Formulation}
Given a search space $S$ of graph neural architecture, a train dataset $D_{train}$, a validation dataset $D_{val}$ and estimation metric $M$, our target is to gain the optimal graph neural architecture $s_{opt} \in S$ which is trained on set $D_{train}$ based on loss function $L$ and achieves the best metric $M^*$ on set $D_{val}$, estimation metric $M$ is represented by accuracy for node classification task. 
Mathematically, it is expressed as follows:

\begin{equation}
\begin{aligned}
 s_{opt}&=argmax_{(s\in S)} M(s(\theta^{*}),D_{val}) \\
 \theta^{*}&=argmin_{\theta } \ L(f(\theta ),D_{train}).
\end{aligned}
\label{eq1} 
\end{equation}

The Figure \ref{fram} shows the parallel search framework, which applied to solve Eq. \ref{eq1}.

\subsection{Search Space}
It is flexible to customize the search space of graph neural architecture. 
Typically, one layer of GNNs involves five architecture components:

\textbf{Attention function $(att)$}: Although each node need to aggregate information from its neighbors, different neighbor node has different contribution for the central node \cite{you2019position}.  
The attention function aims to compute the correlation coefficient $r_{ij}$ for each edge relationship $e_{ij}$ between the nodes $v_i$ and $v_j$.
The types of attention functions are shown in Table \ref{tab:1}.

\textbf{Aggregation function $(agg)$}: When we get all correlation coefficient $r_{ij}$ for the central node $v_i$ in its receptive field $N(v_i)$, in order to achieve central node hidden representation $h_i$, the aggregation operator is required to merge the feature information from the neighbor nodes based on correlation coefficient $r_{ij}$ \cite{hamilton2017inductive}. 
The aggregation manners include $mean$, $max$ and $sum$ in this work.

\textbf{Activation function $(act)$}: After acquiring the node hidden representation $h_i$, for the sake of smoothing $h_i$, a non-linear transformer is usually used.
The choice of activation functions in our $GraphPAS$ is listed as follows: $ \{tanh, sigmoid, relu, linear, softplus, \newline leaky\_relu, relu6, elu \}$.

\textbf{Multi-head $(head)$}: Instead of using the attention function once, research \cite{velivckovic2017graph} shows that multiple independent attention operators can stabilize the learning process.
We collect the number of multi heads within the set: $\left \{1, 2, 4, 6, 8 \right \}$.

\textbf{Hidden dimension $(dim)$}: The node hidden representation $h_i$ is multiplied by a trainable matrix $W^{(l)}$in the each output layer of GNNs, $l$ is the number of GNNs layer, which is used to enhance the hidden representation and reduce feature dimensions.
In this work, the set of hidden dimensions is $\left \{8, 16, 32, 64, 128, 256, 512 \right \}$.

\begin{table}
\caption{Attention functions of search space $S$.}
\label{tab:1}
\begin{center}
  \begin{tabular}{c|c}
    \toprule
    Attention Functions & values\cr
    \midrule
    \midrule
    gat & $r_{ij}^{gat}=reaky\_relu(W_c * h_i + W_n * h_j)$ \cr
    gcn & $r_{ij}^{gcn}=1/ \sqrt{d_i d_j}$ \cr
    cos & $r_{ij}^{cos}=<W_c * h_i, W_n * h_j>$ \cr
    const & $r_{ij}^{const}=1$ \cr
    sym-gat & $r_{ij}^{sym}=r_{ij}^{gat}+r_{ji}^{gat}$ \cr
    linear & $r_{ij}^{lin}=tanh(sum(W_c * h_i))$ \cr
    gene-linear & $r_{ij}^{gen}=W_b * tanh(W_c * h_i + W_n * h_j)$ \cr
    \bottomrule
  \end{tabular}
  \end{center}
 \vspace{-2em}
\end{table}

\subsection{Search Algorithm}
With the analyzing of architecture features of GNNs, we find that different architecture component value has different frequency of occurrence in the GNNs architectures that can get good performance on given dataset.
Inspired by the theory that association algorithm mines frequent itemset\cite{hahsler2007introduction}, in order to restrict search direction in the huge search space for finding good GNNs architecture faster, we use architecture information entropy to dynamically evaluate the mutation selection probability. Next, architecture components are selected based on the mutation selection probability to perform random mutation operation.

\textbf{Parallel Initialization}.
We use multiple workers to explore different areas in the huge search space simultaneously for speed up search process.
Using $N$ workers to randomly explore $N\times M$ GNNs architectures as the initial population in parallel, estimating the validation accuracy as the fitness on the dataset $D_{val}$ for GNNs architectures of initial population.

\textbf{Sharing Population}.
The sharing mechanism is designed to make each worker can utilize the GNNs architecture generated by other worker, it can improve diversity of population for evolution search.
Ranking GNNs architectures based on fitness for initial population, and extracting top $n$ GNNs architectures as the sharing population. 

\textbf{Restricting Search Direction}.
Computing information entropy $h(c_i )$ for each architecture component by Eq. \ref{eqution2} based on frequency distribution of architecture component value in sharing population.

\begin{equation}
\begin{aligned}
h({c}_i)=&-\sum_{j}f({v}_j)log_2f({v}_j)\\
&i\in [1,5\times l], l\geqslant 1 ;
\end{aligned}
\label{eqution2} 
\end{equation}

where $h({c}_i)$ represents the information entropy of the $i$-th component of GNNs architecture, ${v}_j$ denotes the $j$-th value of the $i$-th component, which appears in sharing population, counting ${v}_j$ frequency of occurrence as $f({v}_j)$, $l$ is the number of GNNs layer.

Then, acquiring architecture information entropy vector $\vec{H}=[h({c}_1),h({c}_2),...,h({c}_i)]$.
Finally, based on Eq. \ref{eqution3} to get the architecture component mutation selection probability $p_{i}$. And $\vec{P}=[p_{1},p_{2},...,p_{i}]$ is represented as the architecture mutation selection probability vector.

\begin{equation}
\begin{aligned}
p_i= {softmax}(h({c}_i))\\
i\in [1,5\times l], l\geqslant 1.
\end{aligned}
\label{eqution3} 
\end{equation}

The architecture mutation selection probability vector $\vec{P}$ is a soft constraint strategy for evolution search, which can restrict search direction on the region near the GNNs architectures that have a good performance and reserve probability of exploring other areas at the same time. 

\textbf{Parallel Search}.
Using wheel selection strategy to sample $k$ GNNs architectures from sharing population as the parents for each worker. Selecting $m$ architecture components to perform random mutation based on architecture mutation selection probability vector $\vec{P}$. Each worker will generate $k$ child architectures and evaluate fitness for these child architectures on the dataset $D_{val}$.

\textbf{Sharing Children $\&$ Updating}.
Merging $N \times k$ child architectures generated by $N$ workers as the sharing children. Then, putting child architecture into sharing population if the fitness of child architecture is greater than threshold $F$, which $F$ is the average fitness of top $n$ GNNs architectures in sharing population.

\section{EXPERIMENTS}
\label{experiments}
In this section, we conduct extensive experiments on three benchmark datasets to evaluate the effectiveness of $GraphPAS$ for node classification task.
 
\begin{figure*}[htbp]
  \centering
  \includegraphics[width=\linewidth]{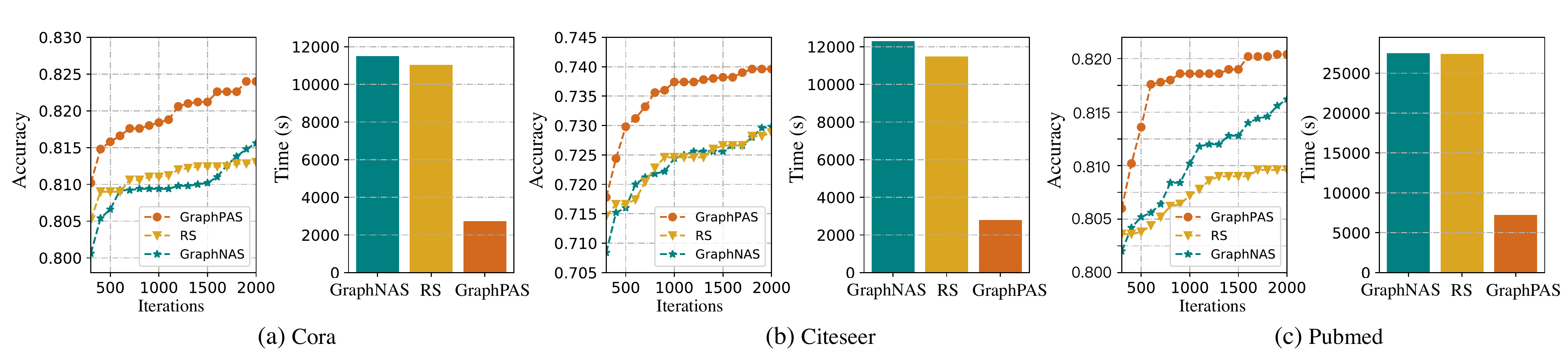}
  \caption{Performance comparison regarding to accuracy and efficiency on dataset Cora, Citeceer, and Pubmed, respectively.}
  \label{framwork}
  \Description{description.}
  \label{efficiency}
\end{figure*}

\subsection{Datasets}
In this work, we use three benchmark citation networks including Citeseer, Cora and Pubmed \cite{sen2008collective} which nodes represent documents and edges denote citation links. 
Dataset characteristics of statistics are summarized in Table \ref{bs2}.

\begin{table}
    \caption{Benchmark Citation Networks.}
    \label{bs2}
    \begin{tabular}{cccc}
        \toprule
        \textbf{Items}& \textbf{Citeseer} &  \textbf{Cora}  &  \textbf{Pubmed}  \\
        \midrule
        \#\textbf{Nodes}            & 3,327    & 2,708 & 19,717  \\
        \#\textbf{Edges}            & 4,732    & 5,429 & 44,338  \\
        \#\textbf{Features}         & 3,703    & 1,433 & 500     \\
        \#\textbf{Classes}          & 6        & 7     & 3       \\
        \#\textbf{Training Nodes}   & 120      & 140   & 60      \\
        \#\textbf{Validation Nodes} & 500      & 500   & 500     \\
        \#\textbf{Testing Nodes}    & 1000     & 1000  & 1000    \\
        \toprule
    \end{tabular}
\vspace{-0.35em}
\end{table}

\subsection{Baseline Methods}
In this study, we compare GNNs architecture identified by our model against baselines including the state-of-art handcrafted architectures and NAS methods for node classification task.
Handcrafted architectures are included: Chebyshev \cite{defferrard2016convolutional}, GCN \cite{kipf2016semi}, GraphSAGE \cite{hamilton2017inductive}, GAT \cite{velivckovic2017graph}, LGCN \cite{gao2018large} and LC-GCN/GAT \cite{xu2020label}. 
NAS methods are involved: Auto-GNN \cite{zhou2019auto}, GraphNAS \cite{gao2020graph} and Gene-GNN \cite{shi2020evolutionary}.

\subsection{Parameter Settings}
The GNNs architecture designed by $GraphPAS$ are implemented by the Pytorch-Geometric library\cite{fey2019fast}.

\textbf{Hyper-parameters of search algorithm:} The parallel workers $N$ as 4, the number of layer $l$ is fixed to 2, the size of random initial population $M$ as 100 for each worker, the number of sharing population $n$ and the number of parents $k$ as 20, the number of mutation $m$ is set to 1, 2, 3, 4 for each worker separately, finally the $search\_epoch$ as 20.

\textbf{Hyper-parameters of GNNs:} For training one GNNs architecture designed by $GraphPAS$, the training $epoch$ as 300, and we used the $L2$ regularization with $\lambda$ = 0.0005, the learning rate $r$ = 0.005, and dropout probability $p$ = 0.6, we utilized $Adam$ $SGD$ to minimize cross-entropy loss function.

\subsection{Test Performance Comparison}
To validate the performance of our model, we compare the GNNs models discovered by $GraphPAS$ with handcrafted models and those designed by other search approaches.
Test accuracy on the node classification task is summarized in Table \ref{result}, where the best performance is highlighted in bold.

\begin{table}[!htbp]
\caption{Test performance comparison of node classification.}
\label{result}
\resizebox{\linewidth}{21mm}{
    \begin{tabular}{ccccc} \toprule
    \textbf{Category} & \textbf{model} & \textbf{Cora} & \textbf{Citeseer} & \textbf{Pubmed} \\ 
    \toprule
    \multirow{6}{*}{\shortstack{\textbf{Handcrafted} \\ \textbf{Architectures}}}
    & Chebyshev & 81.2\% & 68.8\% & 74.4\%  \\
    & GCN & 81.5\% & 70.3\% & 79.0\% \\
    & GAT & 83.0$\pm$0.7\% & 72.5$\pm$0.7\% & 79.0$\pm$0.3\% \\
    & LGCN & 83.3$\pm$0.5\% & 73.0$\pm$0.6\% & 79.5$\pm$0.2\% \\
    & LC-GCN & 82.9$\pm$0.4\%  & 72.3$\pm$0.8\% & 80.1$\pm$0.4\% \\
    & LC-GAT & 83.5$\pm$0.4\%  & 73.8$\pm$0.7\% & 79.1$\pm$0.5\% \\
    \hline
    \multirow{4}{*}{\shortstack{\textbf{NAS} \\ \textbf{Method}}}
    & Auto-GNN    & 83.6$\pm$0.3\% & 73.8$\pm$0.7\% & 79.7$\pm$0.4\% \\
    & GraphNAS    & 83.$7\pm$0.4\% & 73.5$\pm$0.3\% & 80.5$\pm$0.3\% \\
    & Gene-GNN & 83.8$\pm$0.5\% & 73.5$\pm$0.8\% & 79.2$\pm$0.6\% \\
    & \textbf{GraphPAS}    & \textbf{83.9$\pm$0.4\%} & \textbf{73.9$\pm$0.3\%} & \textbf{80.6$\pm$0.3\%} \\ 
    \toprule
    \end{tabular}}
\end{table}

Considering the deviation of testing accuracy, the test performance of $GraphPAS$ is averaged via randomly initializing the optimal GNNs architecture 20 times, and those of handcrafted architectures as well as NAS methods are reported directly from their works.
In general, the NAS methods can achieve better results than handcrafted methods, which demonstrated that it is effective for NAS methods to design good GNNs architecture for the given graph data. 
Especially, the $GraphPAS$ get the best performance on all three benchmark datasets. 
The reason is that the handcrafted models need a lot of tuning of GNNs architecture parameters by manual method for different graph datasets, which is hard to obtain an optimal model. 
On the contrary, the NAS methods can automatically search the GNNs architecture with the ability that makes the performance of GNNs model better for given graph datasets, which can gradually optimize the GNNs model performance with litter human intervention. 

\subsection{Search Efficiency Comparison}
To further investigate the effectiveness of the $GraphPAS$, we compare the progression of top-10 averaged validation performance of GraphNAS, random search and $GraphPAS$.

For each search approach, 2,000 GNNs architectures are explored in the same search space.
As shown in Figure \ref{efficiency}, $GraphPAS$ is more efficient to find the well-performed GNNs architectures during the search process.
Because $GraphPAS$ can search the GNNs architecture in different areas simultaneously in the huge search space based on parallel ability.
Then, the sharing mechanism can enhance the diversity of population, which is beneficial to improve the search efficiency for evolution algorithm.
Finally, $GraphPAS$ utilizes architecture information entropy to dynamically compute mutation selection probability that can restrict search direction effectively for achieving good GNNs architecture faster. 

\section{CONCLUSION}
In this paper, we propose a parallel architecture search framework $GraphPAS$ for graph neural networks. $GraphPAS$ utilizes parallel ability and restricts search direction effectively to speed up search process and achieve optimal GNNs architecture automatically. Experiment results on the three benchmark datasets show that $GraphPAS$ can obtain GNNs architecture that outperforms the state-of-art method with efficiency and accuracy. 

\section{ACKNOWLEDGMENTS}
This work was supported by the National Natural Science Foundation of China (61873288), Zhejiang Lab (2019KE0AB01), Hunan Key Laboratory for Internet of Things in Electricity (2019TP1016), and CAAI-Huawei MindSpore Open Fund.

\bibliographystyle{unsrt}
\balance 
\bibliography{reference}
\end{document}